\documentclass{article}

\usepackage{amsmath,amsfonts,amssymb,times,graphicx,natbib,algorithm,algorithmic}

\usepackage[accepted]{whi2016}

\newtheorem{proposition}{Proposition}

\newcommand{\citeasnoun}[1]{\citeauthor{#1} (\citeyear{#1})}

\newcommand{\Lt}{\mathcal{L}_{\mathcal{T}}}
\newcommand{\Ls}{\mathcal{L}_{\mathcal{S}}}

\icmltitlerunning{Interpretable Recommender via Loss-Preserving Transformation}

\begin{document} 

\twocolumn[
\icmltitle{Building an Interpretable Recommender via Loss-Preserving Transformation}

\icmlauthor{Amit Dhurandhar}{adhuran@us.ibm.com}
\icmlauthor{Sechan Oh}{seoh@us.ibm.com}
\icmlauthor{Marek Petrik}{mpetrik@us.ibm.com}
\icmladdress{IBM T.J. Watson Research Center,
            1101 Kitchawan Road, Yorktown Heights, NY, 10598 USA}


\vskip 0.3in
]
\begin{abstract} 
We propose a method for building an interpretable recommender system for personalizing online content and promotions. Historical data available for the system consists of customer features, provided content (promotions), and user responses. Unlike in a standard multi-class classification setting, misclassification costs depend on both recommended actions and customers. Our method transforms such a data set to a new set which can be used with standard interpretable multi-class classification algorithms. The transformation has the desirable property that minimizing the standard misclassification penalty in this new space is equivalent to minimizing the custom cost function. 
\end{abstract} 

\section{Introduction}

Predictive analytics has been widely used to support decision making in various applications such as online marketing, health care, and personalized recommender systems. In these applications, interpretability of the recommendation logic is critical for the adoption of the decision support tool. For this reason, there has recently been increased focus on interpretable machine learning methods, e.g.~\cite{malioutov2013exact,kim2015MGM,ide2015}. 

Our work is motivated based on a growing need for a (commercial) system that recommends personalized content and promotions in an explicable way. Many marketing managers, who are the main users of the tool, are uncomfortable with fully relying on ``black-box'' recommendation algorithms that cannot be understood, verified, or adjusted to fit their needs. Thus, the rules need to be simple and interpretable. An example rule can be ``if the customer is younger than 30, and spends more than \$500 per month, then show variant 1.'' The number of rules should be small in order to be reviewed and comprehended by marketing managers. In such a situation, ensuring interpretability as well as maintaining the quality of the recommendation is a challenging problem.  

Historical data available in such recommender systems usually consists of the following triples: (\textit{customer features}, \textit{action}, \textit{outcome}). In the context of personalized recommendation, such data sets are suitable to estimate the conversion probability, i.e., the probability that the customer converts to purchase,  under each possible action. Without the interpretability constraint, the recommender system can simply compare the estimated conversion probabilities for an incoming customer under all possible actions and recommend the option with the highest conversion probability such as in \citeasnoun{domingos1999metacost}. In this case, even when the estimator of the conversion probability under each option is interpretable, the optimal recommendation policy may not be summarized in small number of interpretable rules. 

There are many algorithms for learning interpretable rules in multi-class classification problems---such as decision trees~\cite{Hastie2009}---which can be applied to data consisting of pairs of (\textit{customer features}, \textit{action}). Most such learning algorithms compute a set of rules that (approximately) minimizes the misclassification error, which is uniform over customers and classes. The penalty of an suboptimal recommendation, however, is not uniform and varies significantly with different customers and recommended actions. In this paper, we develop a method that transforms the original data with sample dependent costs into a new data set with an identical standard (or 0/1) misclassification error. With the transformed data, one can use existing multi-class classification algorithms to obtain recommendation rules. 

Our work is closely related to cost-sensitive classification problems~\cite{Ling2008}, which are prevalent in practice. Algorithms have been developed for learning cost-sensitive trees, e.g.~\cite{Drummond2003,Ling2004,Lomax2013}. The advantage of our transformation is its generality as it can be used with any classifier. Several such transformation have been previously proposed and studied for binary and multi-class classification~\cite{zadrozny2003cost,Abe2004,Zhou2010}. \citeasnoun{Abe2004} is closest to our work, but in our case the original data set is different from the space of the transformed data set and we study a particular application to recommender systems. The existing research is concerned with a data set consisting of (\textit{features}, \textit{action}) records, and the action represents the right class to which the feature needs to be classified. In contrast, our research deals with a data set consisting of (\textit{features}, \textit{action}, \textit{outcome}) records. In our case, the actions captured in the training data can even be completely independent of the effectiveness of the action depending on how previous actions were determined, i.e., it does not necessarily represent the correct class.  

The transformation method may be more broadly applicable than just to interpretable rule generation for recommender systems. There is almost always a cost associated with taking a particular action in some context. For instance, in healthcare patients being administered a particular treatment may show different levels of recovery and consequently the benefit/detriment to them may vary. Our proposed method can also be used in these other settings.

\section{Problem definition}
Consider an online recommender system for personalized contents and promotions. For an incoming customer, the system can recommend a content or a promotion (an action) from a set of available options $\mathcal{A}$. Each customer is represented by a feature vector $x \in \mathcal{X}$. When action $a \in \mathcal{A}$ is taken, i.e., when option $a$ is provided, to a customer with feature $x$, the customer converts to purchase with probability $p(x,a)$. The optimal option for a customer with feature $x$ is the maximizer of the conversion probability, which we denote by $a^*(x) = \arg\max_{a \in \mathcal{A}} p(x,a)$. The optimal recommender (classifier) $h(\cdot): \mathcal{X} \rightarrow \mathcal{A}$ maximizes the expected conversion rate $E_{x}[p(x,h(x))]$, where the expectation is taken over the distribution of customer feature $x$. The objective is to obtain a near optimal classifier that consists of a small number of interpretable rules. 

We are given $\mathcal{S} = \{ (x_1,a_1,o_1)), \ldots, (x_N,a_N,o_N)\}$, which consists of historical customer feature $x_n$, taken action $a_n$, and the realized outcome $o_n \in \{0,1\}$. Given this data set, one can build an estimator $f(x,a)$ for the conversion probability $p(x,a)$. Without the intrepretability constraint, the recommender system can simply recommend an action $a$ with the highest estimated conversion rate for the given customer feature, i.e., classify $x$ to $\mbox{argmax}_a f(x,a)$. To obtain interpretable recommendation rules using existing multiclass classification algorithms, we transform the data set $\mathcal{S}$ to a new set $\mathcal{T}$ whose elements are pairs of the customer feature $x$ and an action $a$.  

\section{Loss preserving transformation}
One trivial approach to construct $\mathcal{T}$ is to discard a record $(x_n,a_n,o_n) \in \mathcal{S}$ if $o_n = 0$, and otherwise add $(x_n,a_n)$ to $\mathcal{T}$. This approach is problematic, for example, if the prior actions are not uniformly distributed. If a certain promotion option was heavily used before, then recommender trained with $\mathcal{T}$ will classify most inputs to this option. 

A more appealing approach to build $\mathcal{T}$ is constructing an estimator $f(x,a)$ for the conversion probability, and for each $(x_n,a_n,o_n) \in \mathcal{S}$ putting $(x_n,\mbox{argmax}_{a \in A} f(x_n,a))$ to $\mathcal{T}$. In this case, the two sets have the same size, but the action taken in the past is replaced with an estimated optimal action. For each $(x,a) \in \mathcal{T}$, classifying $x$ to an action other than $a$ incurs some misclassifcation penalty, and thus it is encouraged to classify $x$ to $a$. We use this transformation method as the benchmark method. 

Although every element in $\mathcal{T}$ constructed by the benchmark method contains the estimated optimal action for the given input and thus encourages optimal classification, the approach does not reflect the impact of misclassification cost properly. To see this point, consider a classifier $h(\cdot): \mathcal{X} \rightarrow \mathcal{A}$. For a customer with feature $x$, the cost of the classifier $h(x)$ is $p(x,a^*(x)) - p(x,h(x))$, which is the difference between the optimal conversion rate and the conversion rate under the recommended action $h(x)$. Thus, the total loss of the classifier $h(\cdot)$ on the data set $\mathcal{S}$ is 
\[\mathcal{L}_{\mathcal{S}}(h) = \sum_{n=1}^N \left[p(x_n,a^*(x_n)) - p(x_n,h(x_n))\right]~.\]

Now suppose that for some $x_n$, $p(x_n,a)$ is the same for every $a$, i.e., $p(x_n,a) = p(x_n,a^*(x))$ for every $a \in \mathcal{A}$. In this case, there is no cost in recommending any action for $x_n$. Hence, for the purpose of obtaining recommendation rules, $(x_n,a^*(x_n))$ is useless, i.e., it should be effectively removed from $\mathcal{T}$. Next, suppose that for some $x_n,$ $p(x_n,a^*(x_n)) = 1$ and $p(x_n,a) = 0$ for every $a \neq a^*(x_n)$. In this case, classifying $x_n$ to an suboptimal action is always $1$, which is the maximum loss in the conversion rate. Thus, one may want to ensure that $x_n$ is classified to $a^*(x_n)$ in the recommendation rules. These examples imply that misclassification penalties depends on the feature and the classes.

To incorporate the feature and class dependent misclassification penalty, we can control the sample weights (or similarly the number of replicas) in $\mathcal{T}$. Consider the following construction procedure for $\mathcal{T}$: for each $n \in \{1,\ldots,N\}$ and $a \in \mathcal{A}$, set let $k_n^a$ be the weight of sample $(x_n,a)$ in $\mathcal{T}$. Thus, on $\mathcal{T}$ classifying $x_n$ to an action $a$ incurs the total 0/1 misclassification penalty of $\sum_{\hat{a} \in \mathcal{A}\setminus \{a\}} k_n^{\hat{a}}$. The total 0/1 misclassification penalty of a classifier $h(\cdot)$ on $\mathcal{T}$ is given as 
\[\mathcal{L}_{\mathcal{T}}(h) = \sum_{n=1}^N \sum_{\substack{a \in \mathcal{A} : \\ a \neq h^*(x_n)}} k_n^a ~.\]
As it can be readily shown, this condition is satisfied when for every $n$ and $a$,
\begin{equation} \label{eqn:system}
 \sum_{\hat{a}  \in \mathcal{A}\setminus \{a\}} k_n^{\hat{a}} = K \left[ p(x_n,a^*(x_n)) - p(x_n,a) \right] + L, 
\end{equation}
holds for some $K$ and $L$. Sample weights need to be non-negative and $k_n^a \geq 0$. The next proposition, which follows by simple algebraic manipulation, shows bounds the loss due to (approximately) solving the transformed problem.
\begin{proposition}
Let $K>0$ and let $\tilde{h}$ be an approximate minimizer of $\mathcal{L}_{\mathcal{T}}$. Then:
\[ \Lt(\tilde{h}) \le \frac{\Lt(\tilde{h})}{\Lt(h^*)} \Ls(h^*) + \frac{\Lt(\tilde{h})}{\Lt(h^*)} \frac{L}{K}~. \]
And in particular, $\arg\max_{h} \mathcal{L}_{\mathcal{T}}(h) = \arg\max_{h} \mathcal{L}_{\mathcal{S}}(h)$, by setting $\tilde{h} = h^*$.
\end{proposition}
In other words, solving $\Lt$ optimally will give us the optimal solution for $\Ls$ and if the solution is approximate, the quality of the approximation is better for smaller values of $L$. 

The loss function $\mathcal{L}_{\mathcal{T}}$ can be minimized using standard multi-class classification methods and any optimal classifier is also optimal in terms of $\mathcal{L}_{\mathcal{S}}$.

\begin{proposition} \label{prop:optimality}
The optimal solution to \eqref{eqn:system} is given by:
\begin{small}
\begin{equation}\label{eqn:number}
k_n^a = \frac{1}{|\mathcal{A}|-1}\left( K \left( \sum_{\hat{a} \in \mathcal{A}} q_n^{\hat{a}}\right) - K (|\mathcal{A}|-1)q_n^a + L \right),
\end{equation}
\end{small}
where $q_n^a = p(x_n,a^*(x_n)) - p(x_n,a)$.
\end{proposition}
The proof follows, for example, by an immediate application of Sherman-Morrison-Woodbury formula.  In addition, since $\min_{a\in\mathcal{A}} q_n^a = 0$ in Proposition~\ref{prop:optimality}, it can be readily shown that $L \ge 0$.

When using sample replication instead of weights, $k_n^a$ may not be an integer. One solution for this issue is to round $k_n^a$ to the nearest integer. Another approach is to insert $\lfloor k_n^a \rfloor$ replicas of $(x_n,a)$ to $\mathcal{T}$, and add one more replica randomly with probability $k_n^a - \lfloor k_n^a \rfloor$. Both approaches will incur a bias between the two loss functions. However, the impact is be minor if $k_n^a$ are much larger than $1$. For this reason one may want to use a large $K$, but the size of $\mathcal{T}$ increases as $K$ increases. 

In practice, the true conversion probability $p(x,a)$ is not given, and thus needs to be estimated using the historical data. Thus, interpretable rule generation for personalized recommendation can be done in three steps. First, build an estimator for the conversion probability using $\mathcal{S}$. Second, with the estimated conversion probability transform $\mathcal{S}$ to $\mathcal{T}$ based on (\ref{eqn:number}). Third, build a classifier using $\mathcal{T}$. 

\section{Numerical experiments}
We conduct numerical experiments to show the value of the proposed transformation method. We use a data set consisting of three million records of price searches on the website of a transportation company including customer features and estimated conversion probabilities under eight different promotion options. Among the three million records, we used two million records to train two classifiers using the CART algorithm. The first classifier was trained on a set that is constructed via the benchmark method (i.e., a set consists of $(x_n,a^*(x_n))$). The second classifier was trained on a set that is constructed by the proposed transformation method. Because the CART algorithm can incorporate sample weights, we use $k_n^a$ defined in (\ref{eqn:number}) as the sample weight for $(x_n,a)$ instead of adding $k_n^a$ replicas of $(x_n,a)$. By increasing the number of rules to generate (number of leaf nodes in the classification tree), we compute the conversion rate under the two classifiers using the remaining one million records under the assumption that the estimated conversion probabilities are true conversion probabilities. Because actions recommended by the classifiers may not be the same as the actual promotion provided in the historical data set, we cannot test the quality of the classifiers in a truly fair way~\cite{li2011unbiased}. 

\begin{figure}[ht]
    \centering
    \includegraphics[trim=70 70 60 70,clip,width=1\columnwidth]{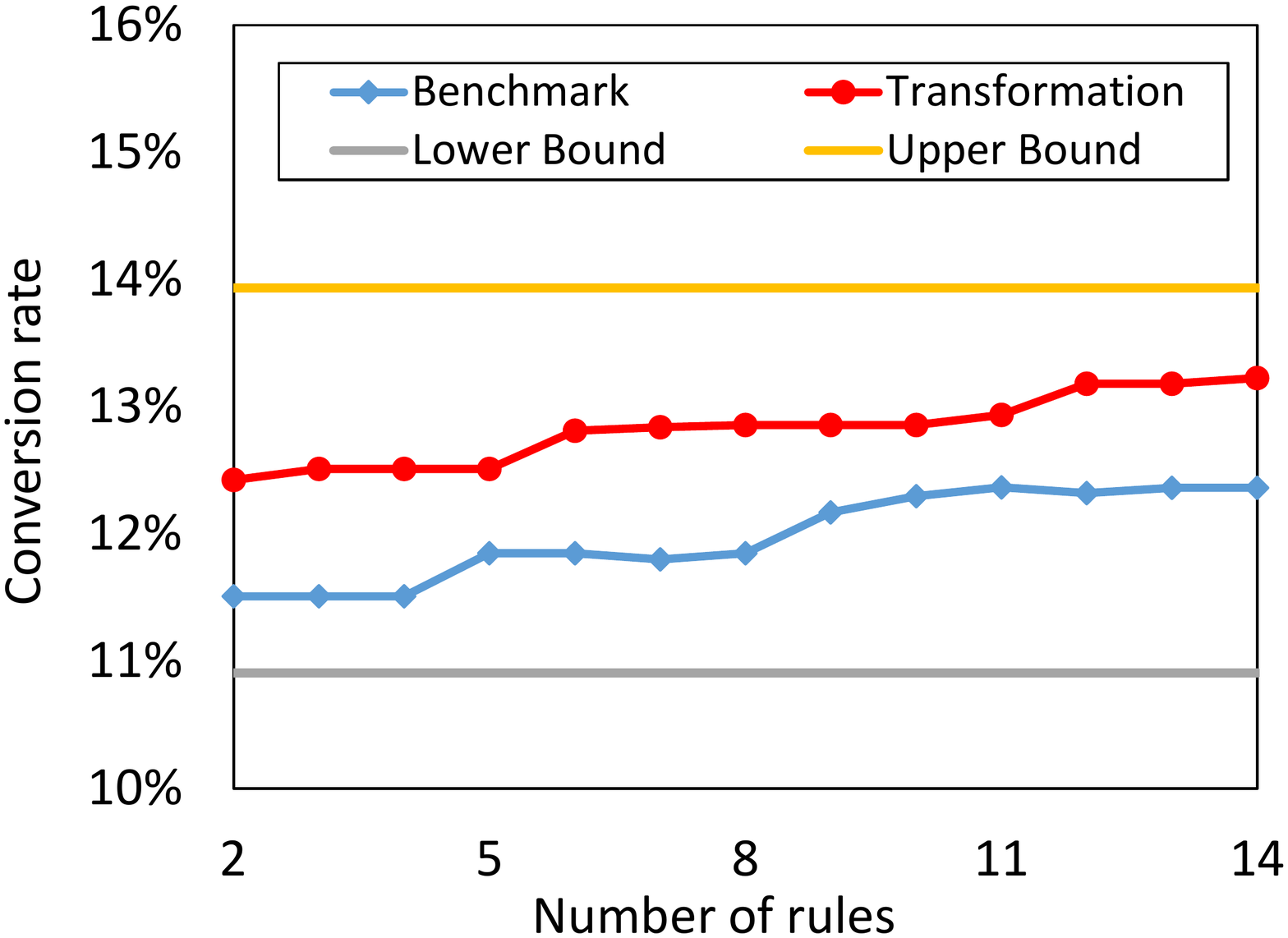}
    \caption{Average Conversion Rates}
    \label{fig:conversion}
\end{figure}
Figure~\ref{fig:conversion} shows the results. The upper bound is the conversion rate when we always recommends the optimal promotion $a^*(x_n)$ for each customer $x_n$, and the lower bound is the conversion rate when we always recommends the worst promotion to each customer (promotion with the lowest conversion probability). The figure shows that for each given number of rules, the recommendation rules obtained via the proposed transformation method have a significantly higher conversion rate than the rules obtained via the benchmark method. Recall that the training data set constructed by the benchmark method contains the exact information on the optimal action for each record. Thus, the deviation from the upper bound may be primarily incurred by the limitation on obtaining a smaller number of interpretable rules. Even under the proposed transformation method, this limitation substantially deteriorates the quality of the recommender. Yet, the deviation from the upper bound is much smaller when we use the proposed conversion method. This result highlights the importance of rigorously addressing the sample and class dependent misclassification loss when obtaining a small number of interpretable classification rules. 

To further show the value of the proposed loss-preserving transformation method, we conducted another numerical experiment. We added one additional fictitious promotion to the original problem. The conversion probability of this promotion is defined as $\max\{ \min_{a \in \mathcal{A}} p(x_n,a), \alpha \max_{a \in \mathcal{A}} p(x_n,a) \}$ for some $\alpha \in [0,1]$. Hence, this promotion performs at least as well as the worst promotion, and performs close to the optimal when $\alpha$ is close to one. Thus, a good classifier would safely recommend this promotion to most customers when $\alpha$ is close to $1$. 
\begin{figure}[ht]
    \centering
    \includegraphics[trim=70 70 60 70,clip,width=1\columnwidth]{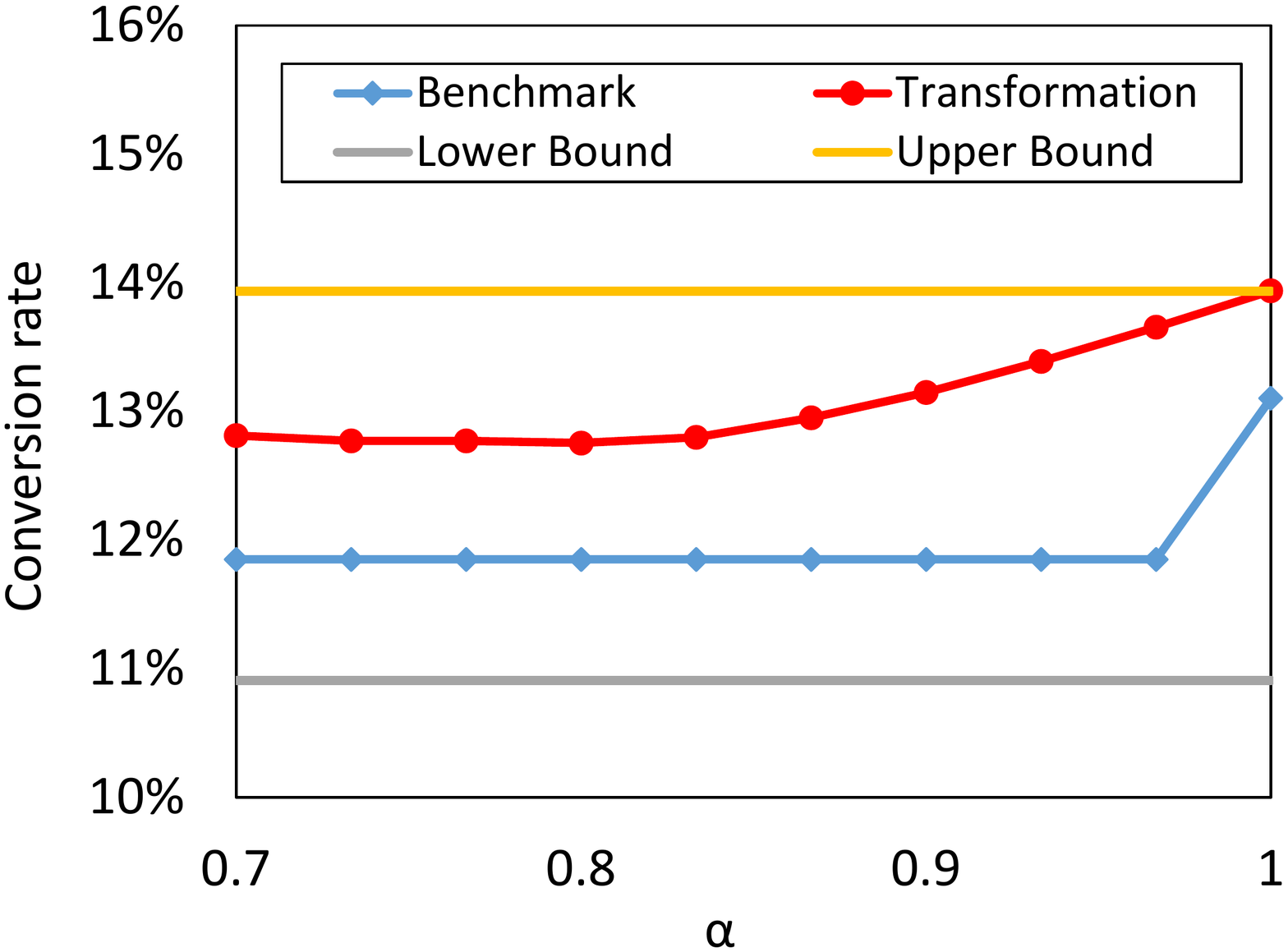}
    \caption{Average Conversion Rates}
    \label{fig:conversion_add}
\end{figure}

We fix the number of rules at six, and repeat the same experiment as before while changing the value of $\alpha$. Note that the fictitious promotion does not change the worst and best conversion probability, and thus the lower and upper bounds do not change. The training set constructed via the benchmark method is not affected by the additional promotion except for when $\alpha = 1$. When $\alpha < 1$, the additional promotion is never optimal, and thus the training data would not have any record containing this additional promotion. Consequently, no rule that recommends this promotion will be obtained. An interesting observation is the fact that the conversion rate under the benchmark method is still notably smaller than the upper bound when $\alpha = 1$. When $\alpha = 1$, always recommending this promotion will achieve the upper bound, i.e., it is globally optimal. However, because there will always be another promotion that achieves the same best conversion, the benchmark method may produce a training data set that contains other promotions, which are only locally optimal. Thus, with the interpretability constraint, the classifier trained out of this data set may fail to produce an optimal set of rules. 

When the proposed transformation method is used, the conversion rate of the recommender increases as $\alpha$ increases in a smooth way, and achieves the upper bound when $\alpha = 1$. The result shows another important reason why rigorously incorporating the sample and class dependent misclassification error is important in interpretable rule generation. The proposed transformation method improves the robustness of classification rules, which often are known to change drastically by a small number of additional data points.   

\section{Discussion and future research}
The proposed transformation method requires conversion probabilities under all actions for each given customer feature, which needs to be estimated from data. The prediction error of the conversion probability estimator will influence construction of the transformed data, which in turn affects the trained classifier. Analysis of the impact of the prediction error in the original space on the quality of the classifier on the transformed space will help improve the transformation method to minimize the true loss of recommendation rules.   

\bibliography{references}

\begin{thebibliography}{13}
\providecommand{\natexlab}[1]{#1}
\providecommand{\url}[1]{\texttt{#1}}
\expandafter\ifx\csname urlstyle\endcsname\relax
  \providecommand{\doi}[1]{doi: #1}\else
  \providecommand{\doi}{doi: \begingroup \urlstyle{rm}\Url}\fi

\bibitem[Abe et~al.(2004)Abe, Zadrozny, and Langford]{Abe2004}
Abe, N., Zadrozny, B., and Langford, J.
\newblock {An Iterative Method for Multi-class Cost-sensitive Learning}.
\newblock In \emph{ACM Conference on Knowledge Discovery and Data Mining
  (KDD)}, pp.\  3--11, 2004.

\bibitem[Domingos(1999)]{domingos1999metacost}
Domingos, P.
\newblock Metacost: A general method for making classifiers cost-sensitive.
\newblock In \emph{Proceedings of the ACM SIGKDD Conference on Knowledge
  Discovery and Data Mining}, pp.\  155--164, 1999.

\bibitem[Drummond et~al.(2003)Drummond, Holte, Chawla, Sheng, Gu, Fang, and
  Wu]{Drummond2003}
Drummond, C., Holte, R.~C., Chawla, Nitesh~V., Sheng, V.~S., Gu, B., Fang, W.,
  and Wu, J.
\newblock {Exploiting the cost (in)sensitivity of decision tree splitting
  criteria}.
\newblock \emph{International Conference on Machine Learning}, pp.\  239--246,
  2003.

\bibitem[Hastie et~al.(2009)Hastie, Tibshirani, and Friedman]{Hastie2009}
Hastie, T., Tibshirani, R., and Friedman, J.
\newblock \emph{{The Elements of Statistical Learning}}.
\newblock Springer, 2nd edition, 2009.

\bibitem[Ide \& Dhurandhar(2015)Ide and Dhurandhar]{ide2015}
Ide, T. and Dhurandhar, A.
\newblock Informative prediction based on ordinal questionnaire data.
\newblock In \emph{Proceedings of IEEE International Conference on Data
  Mining}, pp.\  191--200, 2015.

\bibitem[Kim et~al.(2015)Kim, {Doshi-Velez}, and Shah]{kim2015MGM}
Kim, B., {Doshi-Velez}, F., and Shah, J.~A.
\newblock Mind the {G}ap: {A} generative approach to interpretable feature
  selection and extraction.
\newblock In \emph{Advances in Neural Information Processing Systems}, 2015.

\bibitem[Li et~al.(2011)Li, Chu, Langford, and Wang]{li2011unbiased}
Li, L., Chu, W., Langford, J., and Wang, X.
\newblock Unbiased offline evaluation of contextual-bandit-based news article
  recommendation algorithms.
\newblock In \emph{Proceedings of the ACM International Conference on Web
  Search and Data Mining}, pp.\  297--306, 2011.

\bibitem[Ling \& Sheng(2008)Ling and Sheng]{Ling2008}
Ling, C.~X. and Sheng, V.~S.
\newblock {Cost-sensitive learning and the class imbalance problem}.
\newblock \emph{Encyclopedia of Machine Learning}, pp.\  231--235, 2008.

\bibitem[Ling et~al.(2004)Ling, Yang, Wang, and Zhang]{Ling2004}
Ling, C.~X., Yang, Q., Wang, J., and Zhang, S.
\newblock {Decision trees with minimal costs}.
\newblock In \emph{International Conference on Machine Learning (ICML)}, 2004.

\bibitem[Lomax \& Vadera(2013)Lomax and Vadera]{Lomax2013}
Lomax, S. and Vadera, S.
\newblock A survey of cost-sensitive decision tree induction algorithms.
\newblock \emph{ACM Comput. Surv.}, 45\penalty0 (2):\penalty0 16:1--16:35,
  2013.

\bibitem[Malioutov \& Varshney(2013)Malioutov and Varshney]{malioutov2013exact}
Malioutov, D. and Varshney, K.
\newblock Exact rule learning via boolean compressed sensing.
\newblock In \emph{Proceedings of the International Conference on Machine
  Learning}, pp.\  765--773, 2013.

\bibitem[Zadrozny et~al.(2003)Zadrozny, Langford, and Abe]{zadrozny2003cost}
Zadrozny, B., Langford, J., and Abe, N.
\newblock Cost-sensitive learning by cost-proportionate example weighting.
\newblock In \emph{Proceedings of IEEE International Conference on Data
  Mining}, pp.\  435--442, 2003.

\bibitem[Zhou \& Liu(2006)Zhou and Liu]{Zhou2010}
Zhou, Z. and Liu, X.
\newblock {On Multi-Class Cost-Sensitive Learning}.
\newblock In \emph{AAAI}, pp.\  567--572, 2006.

\end{thebibliography}
\bibliographystyle{icml2016}

\end{document}